\documentclass[journal]{IEEEtran}
\usepackage{amsmath,amsfonts}
\usepackage{algorithmic}
\usepackage{array}
\usepackage[caption=false,font=normalsize,labelfont=sf,textfont=sf]{subfig}
\usepackage{textcomp}
\usepackage{stfloats}
\usepackage{url}
\usepackage{verbatim}
\usepackage{graphicx}
\usepackage{balance}

\usepackage{booktabs}
\usepackage{multirow}
\usepackage{color}

\begin{document}

\title{Normal Transformer: Extracting Surface Geometry from LiDAR Points Enhanced by Visual Semantics}
% \author{Ancheng Lin, Jun Li, ~\IEEEmembership{Member,~IEEE,}}

\author{
Ancheng Lin, Jun Li, Yusheng Xiang, Wei Bian, Mukesh Prasad
% ~\IEEEmembership{Member,~IEEE,}
% <-this % stops a space

\thanks{This work was supported by the China Scholarship Council under Grant 202108200010. (Corresponding author: Jun Li.)}
\thanks{Ancheng Lin, Jun Li and Mukesh Prasad are with the School of Computer Science, Australian Artificial Intelligence Institute (AAII), University of Technology Sydney, Sydney, NSW 2007, Australia (e-mail: ancheng.lin@student.uts.edu.au; jun.li@uts.edu.au; mukesh.prasad@uts.edu.au).} 
\thanks{Yusheng Xiang is with the AI Department of SunnyWay Tech LLC, Suzhou 215000, China (e-mail: yx501182@gmail.com).} 
\thanks{Wei Bian is with Autonavi of Alibaba Group, Beijing 100000, China (e-mail: wei.bian@uts.edu.au).}
}



\maketitle
\maketitle
\makeatletter
\def\ps@IEEEtitlepagestyle{
  \def\@oddfoot{\mycopyrightnotice}
  \def\@evenfoot{}
}
\def\mycopyrightnotice{
  {\footnotesize
  \begin{minipage}{\textwidth}
  \centering
  Accepted by IEEE Transactions on Intelligent Vehicles, with DOI: 10.1109/TIV.2024.3363174
  \\
  Copyright~\copyright~2025 IEEE. Personal use of this material is permitted. However, permission to use this  \\ 
  material for any other purposes must be obtained from the IEEE by sending a request to pubs-permissions@ieee.org.
  \end{minipage}
  }
}

\newcommand{\ve}[1]{{\boldsymbol{#1}}}
\newcommand{\x}{\ve{x}}
\newcommand{\n}{\ve{n}}
\newcommand{\nm}[1]{\n_{#1}}
\newcommand{\set}[1]{{\mathbf{#1}}}
\newcommand{\mat}[1]{{\mathbf{#1}}}
\newcommand{\modelname}{Hybrid Geometry Transformer}
\newcommand{\shortname}{HGT}
\newcommand{\shortnameold}{HGN}
\newcommand{\revised}[1]{{\textcolor{red}{#1}}}
\frenchspacing
\maketitle

\begin{abstract}
High-quality surface normal can help improve geometry estimation in problems faced by autonomous vehicles, such as collision avoidance and occlusion inference. While a considerable volume of literature focuses on densely scanned indoor scenarios, normal estimation during autonomous driving remains an intricate problem due to the sparse, non-uniform, and noisy nature of real-world LiDAR scans. In this paper, we introduce a multi-modal technique that leverages 3D point clouds and 2D colour images obtained from LiDAR and camera sensors for surface normal estimation. We present the Hybrid Geometric Transformer (HGT), a novel transformer-based neural network architecture that proficiently fuses visual semantic and 3D geometric information. Furthermore, we developed an effective learning strategy for the multi-modal data. Experimental results demonstrate the superior effectiveness of our information fusion approach compared to existing methods. It has also been verified that the proposed model can learn from a simulated 3D environment that mimics a traffic scene. The learned geometric knowledge is transferable and can be applied to real-world 3D scenes in the KITTI dataset. Further tasks built upon the estimated normal vectors in the KITTI dataset show that the proposed estimator has an advantage over existing methods.

\end{abstract}


\IEEEpeerreviewmaketitle

\section{Introduction \label{sec:Intro}}

\begin{figure}[htbp]
    \centering
    \includegraphics[width=7.5cm]{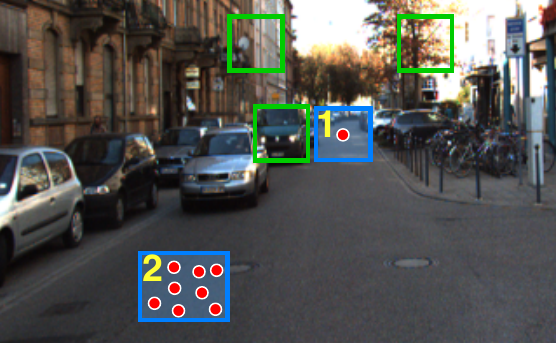}
    \caption{Semantic information in related regions can help estimate surface normals. The figure shows estimation of normal in two regions {\bf R1}: faraway, {\bf R2}: near. R1 is more challenging due to sparse 3D points. The estimation can be assisted by visual information in close/related regions (light green boxes). This figure is best viewed in color.}
    \label{fig:illu}

    \bigskip

    \includegraphics[width=7.5cm]{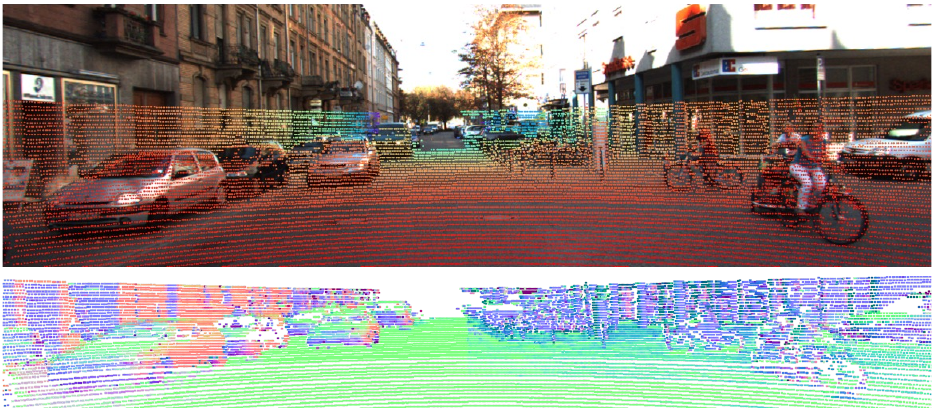}
    \caption{\label{fig:intro_res} Normal estimation on KITTI data. {\bf (top)} LiDAR scans projected to images. {\bf (bottom)} Surface normal estimated by \modelname. }
\end{figure}
\begin{figure*}[htbp]
    \centering
    \includegraphics[width=2.0\columnwidth]{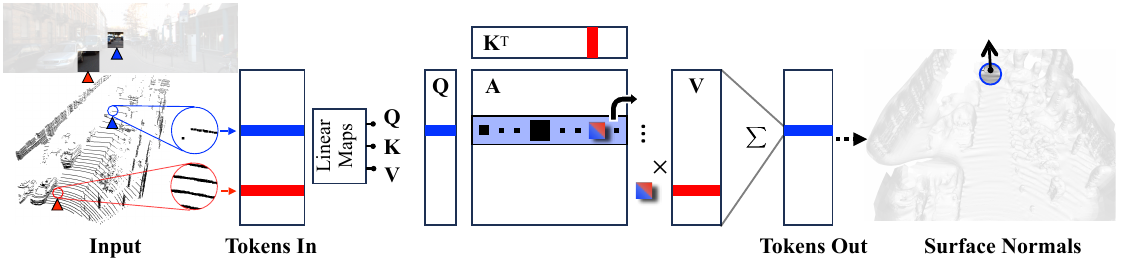}
    \caption{\label{fig:transformer}Mechanism of representing geometric information in an attention block.
\textbf{i)} Tokens from multi-modal inputs are transformed into an association matrix $\mathbf{A}$, where each row determines how a specific position collects information from others. See Section \ref{sec:model} for details. \textbf{ii)} Estimating the normal for a single position (denoted in blue), another position (denoted in red) shows high value in the corresponding row in $\mathbf{A}$.  \textbf{iii)} Although the blue position has sparse neighbours, its normal vector can be accurately estimated with auxiliary information from the red position, which has sufficient neighbours.}
    \label{fig:overview}
\end{figure*}
% ##############################
% # Normal are useful to recover
% ##############################
\IEEEPARstart{D}{ecoding} interesting 3D geometric information of an environment using observations from a specific viewpoint is challenging in autonomous driving. The primary difficulty arises from the inherent insufficiency of information to capture a 3D scene from one viewpoint.
For example, image pixels lack depth information and are subject to perspective ambiguity. On the other hand, individual points in a point cloud contain the distance to the ego-device in 3D space. However, the 3D points are unorganised. The topological and geometrical information of the surfaces is not directly available.

% The neighbourhood of points could help estimate the local geometry of the object surface. 
% , and scene rendering \cite{KovacZ10}
Surface normal estimation has proven valuable across a variety of computer vision and robotic applications such as odometry and mapping \cite{VizzoCCBS21, tiv_GuoZC23}, and 3D reconstruction \cite{KazhdanH13, Huang2023}. In the context of autonomous driving, point clouds with precise normal vectors enrich the understanding of urban environments' spatial relationships. Mid-level surface normal estimation enhances tasks' interpretability and reliability compared to raw point clouds. For instance, decisions for collision avoidance become more comprehensible when based on predicted distances to object surfaces.
% #################################
% # Normal are difficult to recover
% #################################
We consider the problem of estimating the normal vectors of object surface from point clouds. The classical regression-based methods \cite{HoppeDDMS92} formulate surface normal estimation as a least-squares optimisation problem. However, traditional methods could easily become fragile in real scenarios because they are sensitive to noise and manually selected parameters. Recently, learning-based techniques have demonstrated promises to outperform traditional methods \cite{GuerreroKOM18, Zhu21-AdaFit, Li22-GraphFit}. However, these methods are primarily designed for uniform, dense point clouds derived from scanned or synthetic objects. As a result, their applicability to more realistic scenarios, such as those involving LiDAR scans obtained by vehicles, is often limited. 

% ##############################
% # Motivation & "In this work"
% ##############################
In real-life LiDAR scans, points are not uniformly distributed, leading to significant challenges in geometry reconstruction. Specifically, scan points in areas of interest are often scarce, particularly in mid-to-long range fields. As the distance from the device increases, the scan becomes sparser, often resulting in incomplete or missing surfaces. On the other hand, an image provides a high-resolution, dense pixel array representation of the scene, inherently containing rich geometric information. Such images also imply the scene geometry -- biological and machine vision can perform effective geometric inference from images. Therefore, our primary motivation is to leverage this rich semantic information inherent in images to assist in the recovery of surface normal vectors from point clouds.

The normal of a point on a smooth surface is directly determined by a small neighbouring area. Theoretically, one only needs to collect sufficient points of a small region around a point $\x$ to recover the corresponding normal vector $\ve{n}_\x$. However, there can be few measured 3D points close to $\x$. One may turn to image information and fusion-based techniques \cite{KrispelOWPB20}. This is useful when the local texture is sufficient to extract desirable scene information from the observation, but it has limited use for reconstructing detailed geometry of surfaces lacking salient visual cues. To overcome this limitation, regions not necessarily in immediate proximity to $\x$ can also contribute valuable information.

Fig.~\ref{fig:illu} illustrates an example. The masked regions R1 and R2 are both on road surfaces. R2 is closer to the device, containing more LiDAR points. The surface geometry could be directly computed using these points in R2. In contrast, a small neighbourhood may contain only one or a few points in R1. No reliable geometric information can be directly extracted from the small set of scan points. However, if one considers the {\em image} regions marked by green boxes, the geometry in R1 could be implied by the visual information: 
\begin{quote}
    Q: ``What could be a flat surface lying between two buildings and supporting a vehicle beside it?"
    
    A: ``Perhaps a patch of road."
\end{quote}

A naive approach would be adopting convolutional operators to extract local features from the sensory data to represent relevant geometric information, including variants of sparse or graph-based local operators \cite{spconv}. However, determining the scale of a ``local'' area at every point on a manifold is a non-trivial task. And manually crafted procedures also tend to be fragile to deal with complex geometric structures.

To automatically identify and utilise relevant information in the sensory data, this work presents the following insights and innovations:
\begin{itemize}
\item We present a transformer neural network-based model,  \modelname\space (\shortname), to extract geometric information and estimate normal vectors from hybrid data (Fig.~\ref{fig:intro_res}). 
\item We propose an effective training technique for self-attention nets on large-scale outdoor data.
\item We release a data collection toolkit for the Urban environment built on the Unity 3D platform. 
\item The evaluation results show that \shortname\space outperforms previous methods. Furthermore, the tests on KITTI benchmark reveal that our method has excellent generalisation.
\end{itemize}

\section{Related Work\label{sec:Related-Work}}

Normal estimation is an essential task in 3D scene understanding. The simplest and best-known approach is to fit a least-squares local plane using principal component analysis (PCA) \cite{HoppeDDMS92}. Although this method is efficient and well-understood, it is sensitive to noise and density variations and tends to smoothen sharp details \cite{MitraNG04}. To address these challenges, some techniques were proposed to achieve more robust estimation, such as Voronoi cells \cite{MerigotOG11}, distance-weighted approaches \cite{PaulyKKG03} and algebraic spheres' fitting \cite{GuennebaudG07}. 3F2N \cite{Rui2020} is a fast and accurate normal estimator performing three filtering operations on an inverse depth or a disparity image. However, both these traditional methods still depend heavily on the hyper-parameters like neighbourhood size. 

Researchers have applied data-driven deep neural networks widely in 3D shape analysis. In particular, we can group existing models for normal estimation into three categories depending on the data sources.

\textbf{Based on Image:} This kind of work takes a single depth/colour image or both as input and then performs the pixel-wise normal estimation. For a single RGB image, Eigen et al. \cite{EigenPF14} designed a three-scale convolution network to produce better results than traditional methods. Bansal et al. \cite{BansalRG16} and Zhang et al.\cite{ZhangSYSLJF17} adopted the skip-connected structure and U-Net structure to improve performance. Recently, GeoNet++ \cite{QiLLUJ22} uses two-stream CNNs to predict both depth and surface normal maps from a single RGB image. Zeng et al. \cite{ZengTHYSCW19} presented a hierarchical fusion scheme for RGB-D data in a deep learning model and reached state-of-the-art performance.

The common challenge for these image-based methods is the low data efficiency, which means the training procedure requires a large amount of labelled surface normals \cite{LiSDHH15}. In contrast, the point-based approach can improve data efficiency due to the explicit use of intrinsic 3D structure in supervised learning.

\textbf{Based on Point Cloud:} Another group of methods perform 3D scene understanding with point cloud only. Those methods usually use PointNet backbone \cite{QiSMG17} or Graph neural networks (GNNs) \cite{WuPCLZY21} to handle unstructured 3D points. Earlier attempts on normal estimation task came from \cite{GuerreroKOM18}. They use PointNet architecture to extract local 3D shape properties from multi-scale patches of a point cloud. RS-CNN \cite{RS-CNN} and CurveNet \cite{CurveNet} are two milestone works that have significantly enhanced shape awareness and robustness in point cloud analysis. RS-CNN devises a convolutional operator that explicitly encodes the geometric relationships between points and their neighbors. CurveNet introduces a curve-based guided walk strategy within the point cloud to effectively group and aggregate local and global features. Nesti-Net \cite{Ben-ShabatLF19} designed an extra module that learns to decide the optimal scale and hence improves performance. IterNet \cite{LenssenOM20} uses graph neural network to model an adaptive anisotropic kernel that iteratively produces point weights for weighted least-squares plane fitting. Refine-Net \cite{Refine-Net} improves initial normal vectors by utilising multi-scale and multi-type features extracted from both the point cloud and the initial normals. Recently, a series of surface fitting based methods \cite{Ben-Shabat20-DeepFit, Zhu21-AdaFit, Li22-GraphFit} achieve state-of-the-art performance by combining traditional methods with deep learning based layers like Cascaded Scale Aggregation in AdaFit \cite{Zhu21-AdaFit} and graph-convolutional layers in GraphFit \cite{Li22-GraphFit}. \cite{Rethinking_normal_23} analyses the approximation error in these geometry-guided methods, suggesting error reduction through aligning normal vectors with the z-axis and estimating the normal error as a residual term. Distinct from the aforementioned regression methods and surface fitting-based approaches, HSurf-Net \cite{li2022hsurf} proposes to learn hyper surfaces within a high-dimensional feature space, thereby enhancing both robustness and accuracy.

It is worth mentioning that the point cloud data used above is mainly gathered from small CAD objects or indoor depth cameras. These point clouds have an even and relatively high density. In outdoor scenes like urban environments, the heavy interference of the passive illumination \cite{FanelloVRKTDI17} usually makes those active depth sensing solutions fail. In distant areas with low resolutions and small triangulation, stereo methods become less accurate. Hence, LiDAR has become the dominating reliable depth solution in the outdoor environment. However, the point cloud gathered from LiDAR is generally sparse, with noise and a nonuniform point density. Due to the limitations of LiDAR data and the complexity in an outdoor environment, approaches suitable for a dense indoor dataset are often incapable of estimating accurate normals for outdoor data. 

\textbf{Based on colour Image and Sparse LiDAR scans:} 
Various works combine image and LiDAR data to complement each other, ensuring reliable perception in challenging environments. The majority of these studies focus on 3D object detection and segmentation tasks \cite{tiv_SamalKSWM22, KrispelOWPB20, tiv_Li2023_multimodal_review, tiv_seg_peizhou}.

Nonetheless, few investigations have focused on estimating normals from images and sparse point clouds. Closely related to this field, DeepLiDAR \cite{QiuCZZLZP19} uses LiDAR projection maps and images as input and feeds them into a CNN-based encoder-decoder for normal estimation. Distinct from prior works, our approach fuses the two data sources at the feature level instead of the raw data. Moreover, we adopt a point-based transformer architecture to maximise the use of 3D geometric priors in the point cloud.

\begin{figure*}[t]
\begin{centering}
\includegraphics[width=0.6\paperwidth]{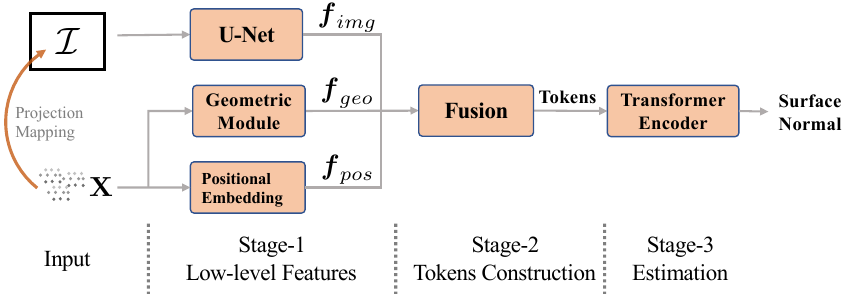}\\
\par\end{centering}
\caption{\label{fig:model} The pipeline of estimating surface normal from image and LiDAR points.}
\end{figure*}

\begin{figure}[t]
\begin{centering}
\includegraphics[width=0.98\columnwidth]{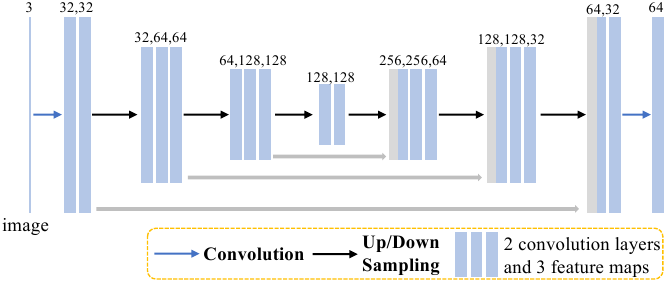}
\par\end{centering}
\caption{\label{fig:unet}Architecture of U-Net \cite{RonnebergerFB15}.}
\end{figure}

\begin{figure}[t] % partitioning point cloud in one frame
    \begin{centering}
    \includegraphics[width=0.35\columnwidth]{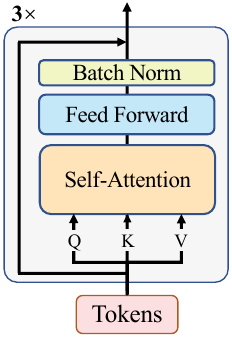}
    \par\end{centering}
    \caption{Transformer encoder with three self-attention blocks.}
    \label{fig:sa_blocks}
\end{figure}

% batch-data
\begin{figure}[t] % partitioning point cloud in one frame
    \begin{centering}
    \includegraphics[width=0.9\columnwidth]{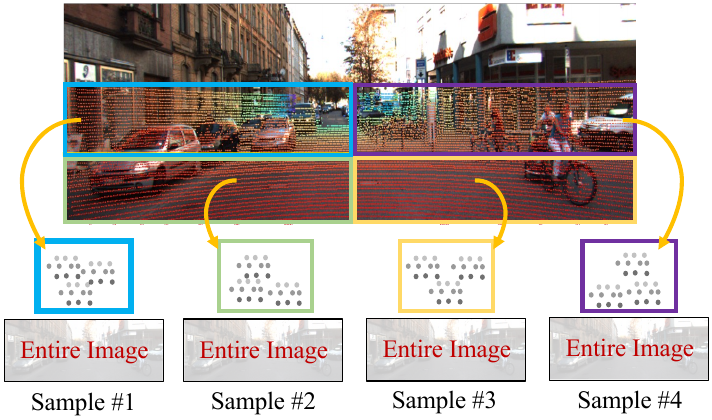}
    \par\end{centering}
    \caption{\label{fig:batch construct} Making small sample by partitioning point clouds.
    The LiDAR points are grouped by the projections on the image into 4 panes. Note the entire image is associated with each group of the points to make the input sample, not only the visual contents in the corresponding pane.}
\end{figure}

% \begin{figure}[t] % partitioning point cloud in one frame
%     \begin{centering}
%     \includegraphics[width=0.9\columnwidth]{fig/transformer_self-attn_block.pdf}
%     \par\end{centering}
%     \caption{\label{fig:transformer} Architecture of transformer-based normal estimator.}
% \end{figure}

\section{Method\label{sec:model}}
This section presents the proposed framework to estimate normal vectors on 3D points from hybrid observation of LiDAR point clouds and images. The computational model is derived from the attention transformers.

Fig.~\ref{fig:model} shows the workflow of the framework.
The input data consist of two parts: an image $\mathcal I$ and a point cloud $\mathbf X$, where $\mathbf X$ is an $[N \times 3]$ array containing the coordinates of a set of 3D points. The point cloud is a subset of raw LiDAR scans, specifically, those points that have a projection onto the image.

The desired output of the model is the surface normal vectors at points in $\mathbf X$. The relative pose of the LiDAR device and the camera is known. Image and geometric features are extracted from $\mathcal I$ and $\mathbf X$, respectively. The features and encoded positions of the points are fed to an attentional transformer (Fig.~\ref{fig:transformer}) to estimate the normal vectors at the points.

The details of the computational steps and the learning of the model are discussed in the remainder of this section.

% MODEL
% - Low-level feature
% - transformer predictor
% LEARN
% - loss 
% - optimisation and batch data preparation
% DATA
% - the synthetic data to help train

\subsection{Low-level features}

As shown in Fig.~\ref{fig:model}, low-level features are descriptors of the 3D points, carrying three pieces of information: local image content, local point cloud geometry and location.

For each 3D point $\ve x_i$, a neighbourhood $\mathcal N_i$ is extracted based on a Euclidean distance. The following image and geometric features at the neighbour points will be aggregated to produce a descriptor $\phi(\ve x_i)$ at $\ve x_i$.

\begin{align}
\phi(\ve x_i) &\leftarrow R(\{\ve{f}_j | j \in \mathcal N_i\}) \label{eq:fusion} \\
\ve{f}_j &\leftarrow  \phi_{fuse}(\ve f_{img}(j) \oplus \ve f_{geo}(j) \oplus \ve f_{pos}(j))  \label{eq:fuse_features}
\end{align}
where the operator $R$ performs a reduction (specifically employing the Max operation) by aggregating the given features. In \eqref{eq:fuse_features}, a feature vector $\ve{f}_j$ is the result of fusing (e.g. using a MLP $\phi_{fuse}$) three-component features: i) image feature $\ve f_{img}$, ii) geometric feature $\ve f_{geo}$ and iii) encoded positions $\ve f_{pos}$. The image and geometric features are derived from widely applied techniques as follows. 

The {\em image features} are computed at each image pixel using a U-Net fully convolutional structure \cite{RonnebergerFB15} as shown in Fig.~\ref{fig:unet}. The U-Net is known for its capability to extract locally semantic information from an image. The hourglass structure enables the net to utilise the information at multiple scales to make predictions. The attribute is desirable for the motivation of utilising semantics to make the normal estimation more robust. An additional advantage is the flexibility regarding input size, making it more generalizable. 

Specifically, the U-Net encodes the raw image $\mathcal I \in \mathbb{R}^{H\times W \times 3}$ into feature map $\ve f_{img} \in \mathbb{R}^{H\times W \times C_{img}}$. Then we obtain $\ve f_{img}(j) \in \mathbb{R}^{C_{img}}$ by taking the feature vector from the 2D position where the 3D point $\ve x_j$ is projected, utilizing the known Camera-LiDAR calibration.

The {\em geometric features} at 3D points are computed using an MLP structure derived from the PointNet++ \cite{QiYSG17}. At each 3D point $\boldsymbol x_j$, $j \in \mathcal N_i$, we first apply a coordinate normalisation
$$
\hat{\x}_j=\x_j-\x_i
$$
and the geometric feature vector is computed by an MLP $\phi_{geo}$: $\mathbb{R}^3 \mapsto \mathbb{R}^{C_{geo}}$ shared by all points.

The {\em location} of a point is encoded by {\em positional embedding} layers $\phi_{pos}$, which employ MLP structures as well. Following the common practice of encoding 2D coordinates in image processing \cite{Vit}, we encode every point's 3D spatial information in the global coordinate system into $\mathbb{R}^{C_{pos}}$.

\subsection{Transformer Encoder and Prediction Head}
The multi-modal feature extraction and aggregation process described in (\ref{eq:fusion}) produces a comprehensive feature vector $\phi(\ve x_i) \in \mathbb R^C$ for each 3D point $\boldsymbol x_i$. All the $N$ points can result in a matrix of tokens $\mathbf T: [N \times C]$. The transformer encoder comprises three self-attention blocks. The first self-attention block takes $\mathbf{T}$ as input and outputs new tokens $\mathbf{T}'$, which then serve as the subsequent self-attention block's input.

Fig.~\ref{fig:sa_blocks} provides a detailed illustration of the pipeline. In each block, $\mat T$ is linearly transformed into three tensors $\mat Q, \mat K \in \mathbb{R}^{N\times D}$, and $\mat V \in \mathbb{R}^{N\times E}$. Then $\mat Q$ and $\mat K$ are used to represent the relationship between the tokens. One commonly adopted computational model of the relation is to take the inner product:
\begin{align}
\mat{A}' &= \left[ a_{i,j} \right]_{N \times N} = \mat{Q}\mat{K}^T  \label{eq:attn_mat} \\
\mat{A} &= \left[ \frac{\hat{a}_{i,j}}{\sum_{k} \hat{a}_{i,k}}\right]_{N \times N}, \quad 
\hat{a}_{i,j} = \frac{\text{exp}(a_{i,j})}{\sum_{k} \text{exp}(a_{k,j})} \label{eq:attn_norm}
\end{align}
where the normalisation operation in \eqref{eq:attn_norm} is following the practice of \cite{PCT}. The entry $\mat{A}_{i,j}$ represents how token $i$ in $\mat Q$ is related to the reference token $j$ in $\mat K$ and $\mat V$.

The model then conducts a matrix multiplication $\mat A \mat V$ for internal feature combinations between tokens. The combined tokens are passed into a Feed Forward layer followed by batch normalisation and residual connection. Finally, the layer outputs new tokens $\mat T'$ serving as the next self-attention block's input.

Following the self-attention blocks, an MLP prediction head $\phi_{pred}$ transforms each encoded token into a 3-dimensional vector representing three components of the surface normal. Finally, we regularise the output to unit 3D vectors to ensure a reasonable surface normal.

\subsection{Loss function}
The loss function for optimising the neural network model is defined as the mean squared error (MSE) between the directions of the estimated and the ground-truth normal vectors. The MSE of the estimations at $N$ points is
\begin{align}
\mathcal{L}=\frac{1}{N}\sum_{j}^{N}\|\ve{n}^{\text{est}}_{(j)}-\ve{n}^{\text{gt}}_{(j)}\|_2^2
\label{eq:loss}
\end{align}
where $\ve{n}^{\text{est}}_{(j)}$ is an estimation by the model and $\ve{n}^{\text{gt}}_{(j)}$ is the corresponding ground-truth normal vector.

\subsection{Effective Training of the Attentional Transformer}
\subsubsection{Batching}
The transformer normal estimator is designed to utilise semantic information across the scene by employing large attentional fields. 
However, the attention model is expensive in terms of computation and storage. 
The attention weight matrix $\mat A \in \mathbb R^{N\times N}$ in \eqref{eq:attn_mat} grows quadratically with the fused features, which is prohibitive for a complex scene.

More specifically, fully executing the model on one typical training sample ($400 \times 400$ image and ~10K points) costs about 6 Gigabytes of GPU memory. The cost is about 4x higher for a single frame in the real traffic scene in the KITTI dataset.
Although straightforward implementation is viable on modern hardware configurations, the sample number in a batch during training must be compromised. At the same time, the batch size can affect the effectiveness of training \cite{how_bn}. 

For effective training, we partition the point cloud in one frame by grouping the points according to their projections on the image. For example, Fig.~\ref{fig:batch construct} (a) shows partitioning the points into 4 parts. The points of each part make one training sample.
Since the input tokens of the transformer net correspond to individual points, the partitioned samples make the size of the weight matrix $\mat A$ decreases quadratically, e.g. from $[N \times N]$ to $[\frac{1}{4} N \times \frac{1}{4} N]$ in the example shown in  Fig.~\ref{fig:batch construct}. Within the storage limit, the batch size can be increased correspondingly, see the right pane of the figure. 
% A batch constructed by this method will not result in an explosion of memory because the matrix elements reduction (from $N^2$ to $4 \times \frac{1}{16} N^2$) saves enough memory to tolerate an x4 increase of image features.
Increased batch is helpful during the training, especially when batch normalisation is employed.

In this paper, we use $\frac{1}{16}$ of each frame's point cloud, and the batch size is 8. The experimental results in Fig.~\ref{fig:att_map} show that our training technique will not lead to a loss of global sensibility during the testing phase. Furthermore, this technique can also be used in the testing phase if the range of scene observation is too extensive.

\subsubsection{Synthetic data}
Measuring normal vectors is difficult in practical applications. And most real-world datasets do not contain annotations of the ground-truth normal vectors. A toolkit is developed to produce synthetic data by simulating a city scene in a realistic 3D renderer environment. Artificial cameras and LiDAR devices are planted in the scene. We construct a customised shader to synthesise the normal vectors at points on object surfaces. 
Data samples are collected in the same form as the actual sensors. More importantly, labels (colours, depth, surface normals) are easily acquired from the simulator. Details about how data is synthesised can be found in the next section. We use this dataset to train and evaluate our model.

Experiments have shown that the model trained only on synthetic data has been able to generalise to practical datasets and obtain impressive results. Further improvement could be achieved if we use real-world data to fine-turn our model.

% Experiment
% - Data Preparation
% - implementation Details
% - Test on Synthetic Data
%   - performance
%        -better than PCPNET; 
%        -better data-efficiency (DeepLidar failed)
%   - Ablation of transformer
%   - visualization of attention map
% - Test on Real-world KITTI
% - Test on Reconstruction Task

\section{Experiments}
This section presents the experimental results of the proposed technique. 
We first introduce how the synthetic data has been constructed. We then evaluate our model on this dataset and compare it with other existing works. We add different levels of noise into the data to test whether our model is robust. The model trained on the synthetic dataset is also applied to the real-life KITTI dataset \cite{GeigerLSU13}. The KITTI dataset does not contain ground-truth normal vectors at the LiDAR points. Hence the estimated normal vectors are used for 3D reconstruction to assess the estimation.

\subsection{Data Synthesis \label{sec:data}}
% \begin{figure}[t]
% \begin{centering}
% \includegraphics[width=0.8\columnwidth]{fig/location-path.pdf}
% \par\end{centering}
% \caption{\label{fig:sensors-loc} Scene and movement setup for data collection.}
% \end{figure}

% \begin{figure}[t]
% \begin{centering}
% \includegraphics[width=0.85\columnwidth]{fig/shaders.pdf}
% \par\end{centering}
% \caption{\label{fig:shaders}Observations rendered by shaders in the 3D rendering engine.}
% \end{figure}

% \begin{figure}[t]
% \begin{centering}
% \includegraphics[width=0.83\columnwidth]{fig/LiDAR-projection-inverse.pdf}
% \par\end{centering}
% \caption{\label{fig:lidar_projection}LiDAR points' projection of KITTI and our synthetic data.}
% \end{figure}

\begin{figure}[t]
\begin{centering}
\includegraphics[width=0.8\columnwidth]{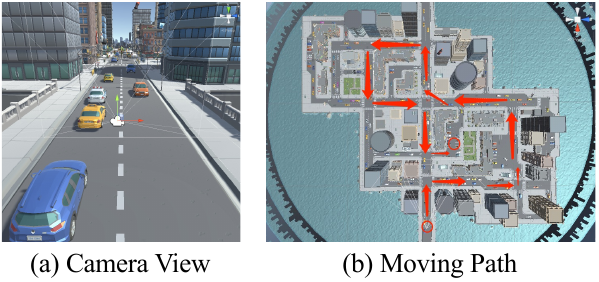}
\par\end{centering}
\caption{\label{fig:sensors-loc} Scene and movement setup for data collection.}

\bigskip

\begin{centering}
\includegraphics[width=0.8\columnwidth]{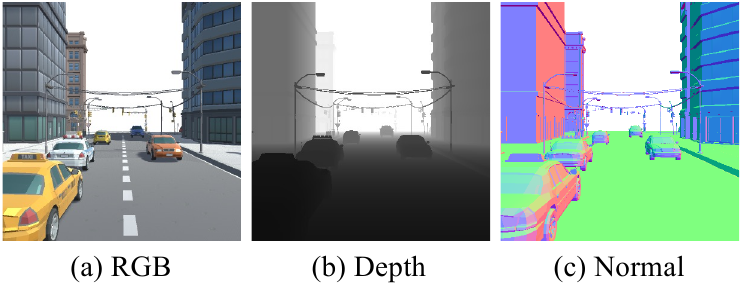}
\par\end{centering}
\caption{\label{fig:shaders}Observations rendered by shaders in the 3D rendering engine.}

\bigskip

\begin{centering}
\includegraphics[width=0.8\columnwidth]{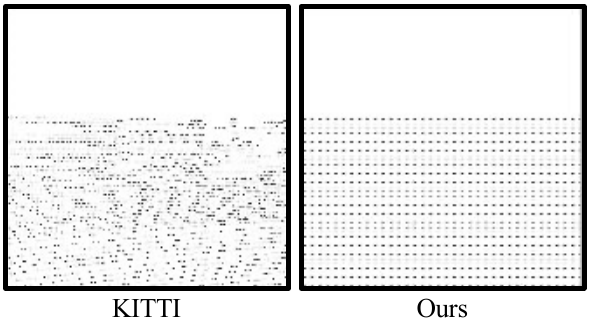}
\par\end{centering}
\caption{\label{fig:lidar_projection}LiDAR points' projection of KITTI and our synthetic data.}

\end{figure}

We exploit Unity 3D \cite{haas2014history} as a simulator and rendering engine to collect data. We first build a high-quality outdoor scene based on the package ``POLYGON CITY"\footnote{https://assetstore.unity.com/packages/3d/environments/urban/polygon-city-low-poly-3d-art-by-synty-95214} on Unity 3D Assert Store. It contains vehicles, cyclists, pedestrians, and necessary lighting configurations. The scene can be easily changed to others according to practical needs. Our efforts mainly focus on the data collection after setting up a scene. More specifically, we develop in the following aspects.

\textbf{Sensors and Shaders:} The sensors we place in the simulator include a colour camera, depth camera and normal sensor. The depth sensor is utilised to produce LiDAR points in the latter process. As shown in Fig.~\ref{fig:sensors-loc} (a), we place these sensors on the roof of the car. We can render the sensor observations by developing engine shaders for the three sensors, as shown in Fig.~\ref{fig:shaders}.

\textbf{LiDAR acquisition:} We find that the LiDAR projections to the image are mostly uniform. Therefore we take the lower rectangular region of the image and 
evenly sample rays passing through the image pixels to simulate projected LiDAR points. Note the points are distributed evenly on the image, but not in the 3D scene. The density has been made comparable to that in the KITTI dataset.
Fig.~\ref{fig:lidar_projection} compares the sampling of the LiDAR points between the synthetic and KITTI datasets.

We then apply the designed mask to the given depth map from the sensor. Finally, the synthetic LiDAR scans can be obtained via inverse projection using camera intrinsic. Unlike other simulators, which spend a lot of time and computational cost on simulating real LiDAR using ray tracing, we simulate LiDAR directly from the depth map.

\textbf{Dataset generation:} With the communication interfaces provided by ML-Agent \cite{ml-agent}, we deploy scripts in Unity and an external Python-based controller, respectively.

We follow the preset path shown in Fig.~\ref{fig:sensors-loc} (b) and collect $151$ frames' observations ($121$ for training, $30$ for testing), which contain images with size $400\times400$, sparse LiDAR points in view, and points' surface normal ground truth. For convenience, we export the projection relationship arrays between LiDAR points and pixel index and the camera projection matrix.

To evaluate the robustness of the model, we also add different levels of noises to the LiDAR data. Note that we only add a numerical drift to the $z$-coordinate (in front of the camera) of the LiDAR point cloud. The noises follow the settings in \cite{Ben-ShabatLF19,GuerreroKOM18}, see Fig.~\ref{fig:res_unity_err}.

\subsection{Implementation Details}

We implement our model using PyTorch \cite{PyTorch} on a NVIDIA RTX 5000 GPU. We adopts Adam \cite{KingmaB14} as optimiser with learning rate $1e^{-4}$, $\beta_1=0.9$, $\beta_2=0.999$. All models are trained for $200$ epochs from randomly initialised parameters.

In the low-level features extraction, for each point, we randomly select $60$ neighbours, and the radius of the spherical query is $0.75$. If there are fewer points within the sphere, we will pad the results with the querying point itself. We have implemented a lightweight network architecture to enhance computational and storage efficiency. Specifically, we utilize a 2-layer MLP for both geometric feature extraction $\phi_{geo}$, positional embedding $\phi_{pos}$, multi-modal feature fusion $\phi_{fuse}$, and prediction head $\phi_{pred}$. The transformer encoder in our model comprises three self-attention blocks.

\subsection{Test on Synthetic Data \label{sec:syn_data}}

\begin{figure*}[t]
\begin{centering}
\includegraphics[width=0.83\paperwidth]{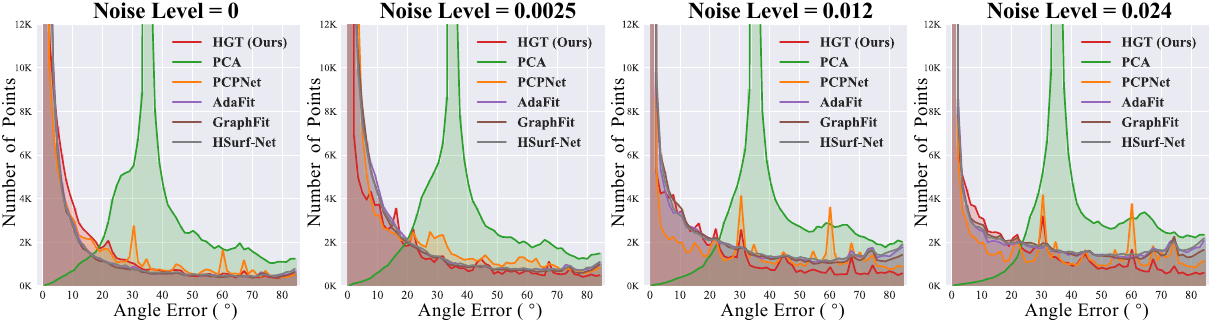}
\par\end{centering}

\caption{\label{fig:res_unity_err}Frequency polygon of errors on the synthetic dataset. Each pane corresponds to varying noise levels. The horizontal axis denotes the range of errors, while the vertical axis indicates the number of points with specific estimation errors. The vertical axis is truncated between 0-12K to emphasize the `long tail' effect, which differentiates the performance of various methods.
}
\end{figure*}

\begin{figure*}[t]
\begin{centering}
\includegraphics[width=0.83\paperwidth]{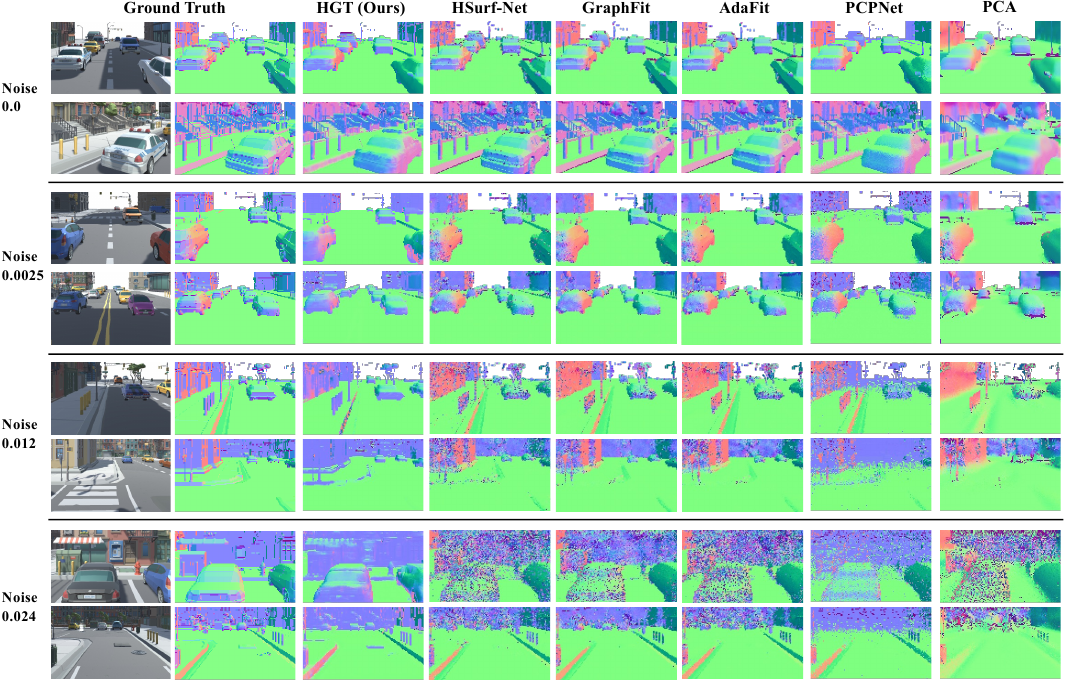}
\par\end{centering}
\caption{\label{fig:res_unity_im} Qualitative comparison on synthetic dataset.}
\end{figure*}

\begin{table}
    \centering
    \caption{Average angle error on synthetic dataset. This table also includes an ablation study showing results when two key components are removed from our final HGT model.}

    \begin{tabular}{lllll}
    \toprule
    \multicolumn{1}{l}{\multirow{2}{1.7cm}{\textbf{Method}}} & \multicolumn{4}{c}{Noise Level}  \\ 
    \cline{2-5}
           & 0       & 0.0025   & 0.012    & 0.024    \\
    \midrule
    PCA \cite{HoppeDDMS92}    & 39.27   & 40.10    & 42.80    & 44.06          \\
    PCPNet \cite{GuerreroKOM18} & 10.50   & 15.94    & 19.50    & 20.83          \\
    AdaFit \cite{Zhu21-AdaFit} & 7.20   & 11.12    & 16.07    & 17.89          \\
    GraphFit \cite{Li22-GraphFit} & \textbf{6.91}   & 10.41    & 14.90    & 17.01          \\
    HSurf-Net \cite{li2022hsurf} & 7.34  & 11.05  & 16.18  & 18.10 \\
    \midrule
    Ours (w/o image) & 10.56   & 12.54    & 13.60    & 14.67 \\
    Ours (w/o transformer) & 8.49   & 10.42    & 11.31    & 11.41 \\
    Ours (full) & 8.18 & \textbf{9.61} & \textbf{10.36} & \textbf{11.38}\\ 
    \bottomrule
    \end{tabular}
    
    \label{tab:table_error}
\end{table}

\begin{table}
    \centering
    \caption{Comparison of time and space efficiency. This table shows the average time required to process one frame in KITTI dataset.}
    \begin{tabular}{lrr}
    \toprule
    \textbf{Method} & Time & \#.params  \\
    \midrule
    PCPNet \cite{GuerreroKOM18}     & 354ms    & 3.47M          \\
    AdaFit \cite{Zhu21-AdaFit}      & 1167ms   & 3.53M          \\
    GraphFit \cite{Li22-GraphFit}   & 6401ms   & 4.26M          \\
    HSurf-Net \cite{li2022hsurf}    & 1231ms   & 2.16M          \\
    \midrule
    HGT (Ours)                      & \textbf{86ms}              & \textbf{1.83M}  \\ 
    \bottomrule
    \end{tabular}
    
    \label{tab:table_efficiency}
\end{table}

\subsubsection{Evaluation Metric}
To compare two directions, we consider directly measure the angles between the vectors. The following trigonometric evaluation is computed.
% \ve{n}^{\text{est}}
\begin{equation}
d_{\text{angle}}=\frac{1}{N}\sum_{j}^{N}\arccos(\frac{\lvert\ve{n}^{\text{est}}_{(j)}\cdot\ve{n}^{\text{gt}}_{(j)}\rvert}{\lvert\ve{n}^{\text{est}}_{(j)}\rvert\lvert\ve{n}^{\text{gt}}_{(j)}\rvert})  \label{eq:angle_diff}
\end{equation}
where $\ve{n}^{\text{est}}_{(j)}$ is an estimation by the model and $\ve{n}^{\text{gt}}_{(j)}$ is the corresponding ground-truth normal vector. This angle difference is consistent with MSE in \eqref{eq:loss}. 

\subsubsection{Performance}
We compare our method with other baseline methods \cite{GuerreroKOM18, Zhu21-AdaFit, Li22-GraphFit, li2022hsurf}, each trained using the same train/test data (121 frames for training, 30 frames for testing; 4 noise levels). Additional settings including the number of training epochs (200) and the number of points in local patches (60), are aligned with those used in HGT to ensure a fair comparison.

Tab.~\ref{tab:table_error} lists the average angular errors for a numerical comparison. In zero-noise data, the performance of \shortname\space is comparable to that of state-of-the-art methods, with a minor deviation of less than 1.5°. The superiority of \shortname\space becomes significant as noise levels escalate. This distinction in robustness will impact performance in real-world urban data, as demonstrated in subsequent tests.

In Fig. \ref{fig:res_unity_err}, we analyse the distribution of estimation errors across all LiDAR points. Specifically, we segment the error range (0-90°) into 60 bins and count the number of points falling within each bin. The distributions reveal that as noise levels increase, the estimations of other methods shift towards the tail (indicating a high error), whereas our method maintains a relatively low and flat tail. Fig.~\ref{fig:res_unity_im} further visualises some of the estimated surface normal maps for qualitative comparison.

The report does not include the results of DeepLiDAR \cite{QiuCZZLZP19} since its pure CNN-Based architecture can not converges well on our small dataset (only $121$ images and point clouds). This also reveals that our hybrid structure is more effective in handling multi-modal data.

\subsection{Generalisation Test on KITTI}

\begin{figure*}
    \begin{centering}
    \includegraphics[width=0.78\paperwidth]{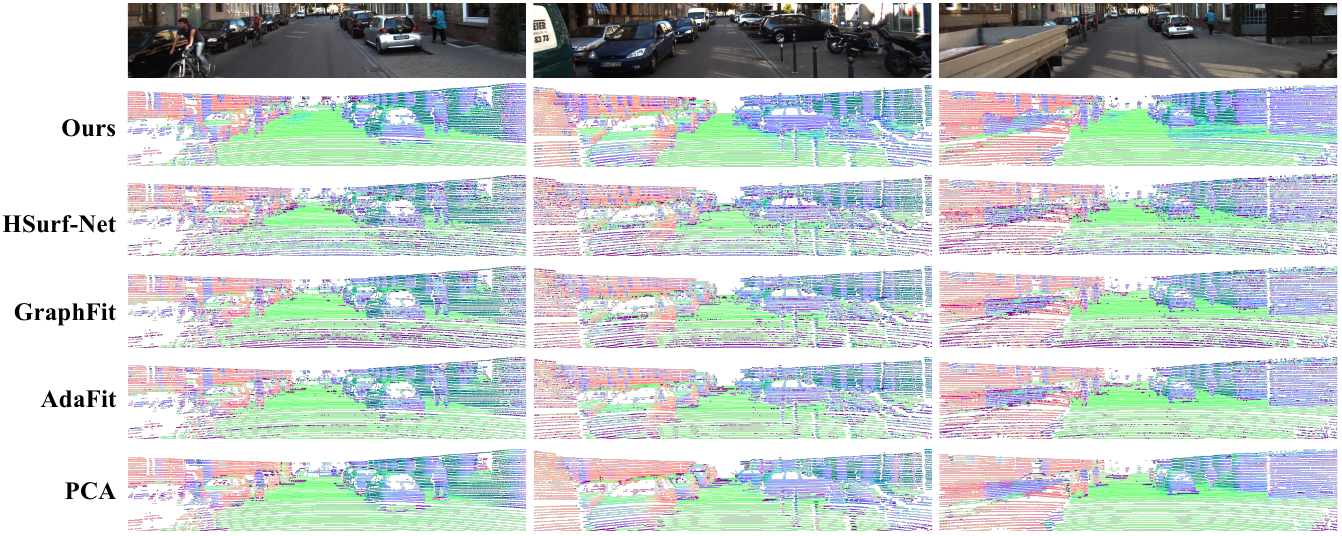}
    \par\end{centering}
    \caption{\label{fig:res_kitti_im}Qualitative results on the KITTI dataset.}
\end{figure*}

\begin{figure*}
    \begin{centering}
    \includegraphics[width=0.8\paperwidth]{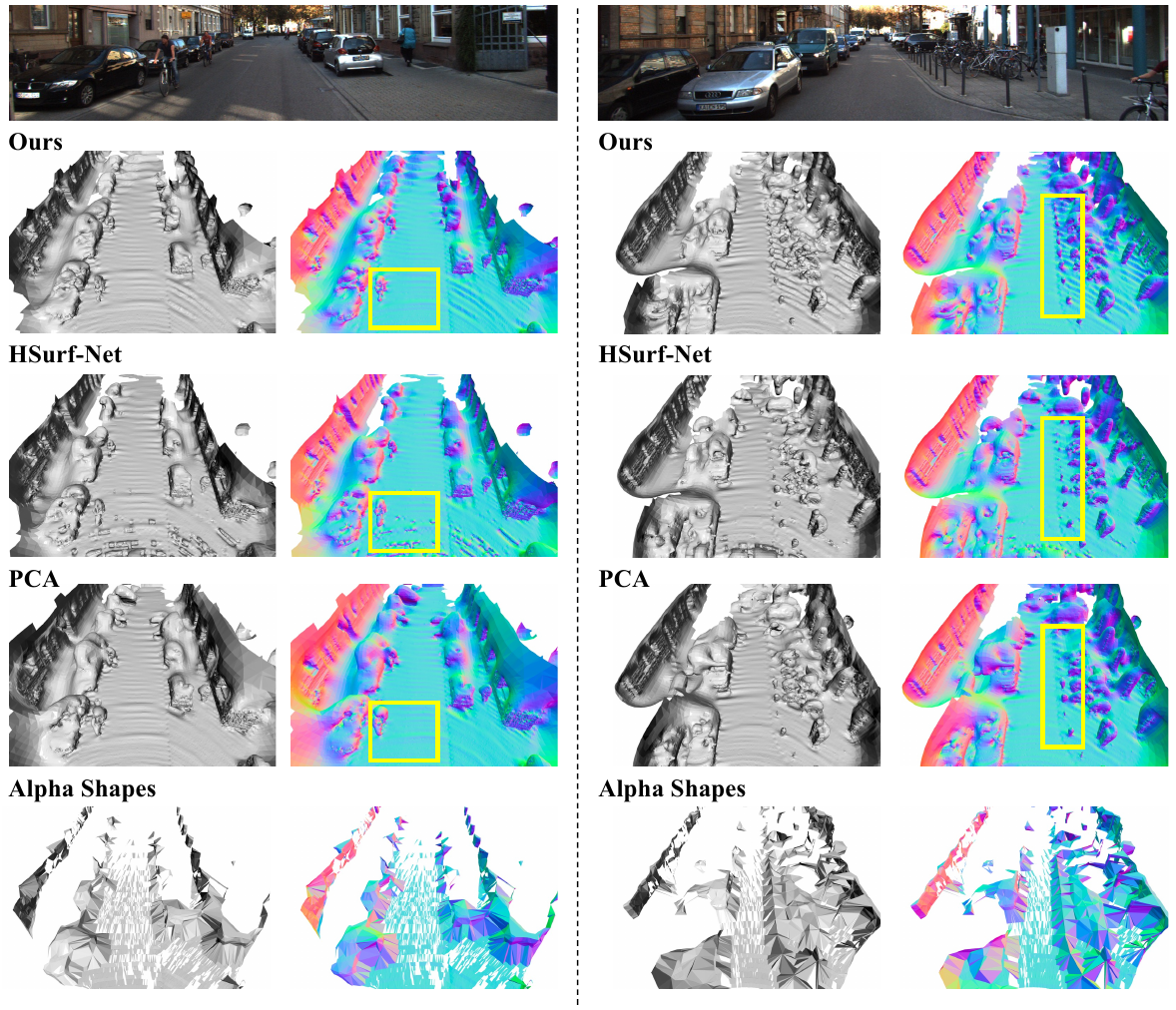}
    \par\end{centering}
    \caption{\label{fig:reconstruction_kitti} Evaluation on 3D reconstruction task. Alpha Shapes represents a point-only reconstruction method, while the others perform Screened Poisson Surface Reconstruction (SPSR) \cite{KazhdanH13} utilizing their respective estimated surface normals.}
\end{figure*}

In this experiment, we use the real-world dataset KITTI \cite{GeigerLSU13} to investigate the model generalisation. Given a frame of LiDAR scans and camera observations of KITTI, we first compute and save the point-to-pixel projection mapping with calibration parameters provided by the dataset. Then we directly feed the KITTI frames into \shortname\space for surface normal estimation.

For comparison, we also illustrate the surface normal maps produced by two recent methods and the classical PCA as in Tab.~\ref{tab:table_error}. Both methods are trained on the same synthetic dataset from Section \ref{sec:syn_data} without additional fine-tuning.

The results are shown in Fig.~\ref{fig:res_kitti_im}. \shortname\space has generalised surprisingly well, given that no fine-tuning has been conducted on the KITTI dataset. Compare the estimated normal vectors of the road surface in Fig.~\ref{fig:res_unity_im} for a specific example.

\subsection{Application on 3D Reconstruction}

Given the estimated normal vectors at the one-frame LiDAR points from the KITTI dataset, we reconstruct the 3D scene using Screened Poisson Surface Reconstruction (SPSR) \cite{KazhdanH13}. The 3D reconstruction could serve as an intuitive criterion to compare the normal estimation methods. We also include comparisons to reconstruction w/o normals.

As shown in Fig.~\ref{fig:reconstruction_kitti}, the point-only method can lead to many mesh holes and is unsuitable for handling practical sparse LiDAR data. The Poisson Reconstruction technique can recover a continuous surface with the help of surface normals. By comparison, we can find that our model outperforms in 3D reconstruction as the quality of the reconstructed model significantly relies on the accuracy of the surface normals.

\subsection{Additional Experiments}

\begin{figure*}
\begin{centering}
\includegraphics[width=1.95\columnwidth]{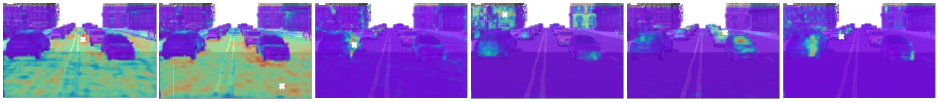}
\par\end{centering}
\caption{\label{fig:att_map} Visualisation of attention weights produced by HGT using the synthetic dataset. White cross markers indicate the positions for normal estimation. Note that we are testing six distinct positions on the same frame.}
\end{figure*}

\begin{figure*}[htbp]
\begin{centering}
\includegraphics[width=1.95\columnwidth]{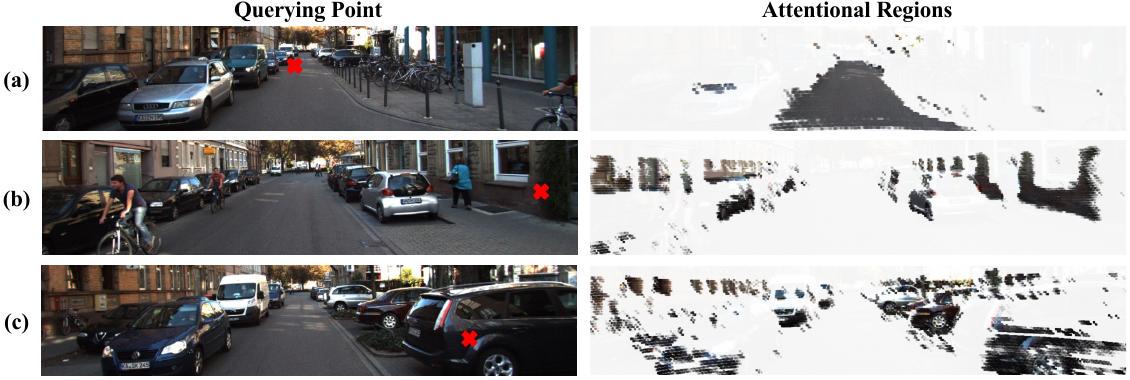}
\par\end{centering}
\caption{\label{fig:att_map_kitti}Visualisation of attention weights produced by HGT using the KITTI dataset. In the left subfigures, red cross markers indicate the positions for normal estimation. In the right subfigures, regions associated with high attention weights are depicted with greater opacity.}
\end{figure*}

\subsubsection{Ablation studies}
To verify the effectiveness of multi-modal information, we train our model using only point cloud data by disabling the U-Net branch. The results in Tab. \ref{tab:table_error} demonstrate that incorporating image input reduces average angle error by approximately 3°.

To investigate how the transformer structure helps to recover the surface normal, we replace self-attention blocks with a PointNet structure. Specifically, an MLP and max pooling operation is applied to tokens and then generates a global feature. We concatenate this global feature to all tokens and pass them to the estimation head. The ablation results can also be found in Tab. \ref{tab:table_error}, revealing that the transformer does bring a performance improvement.

\subsubsection{Efficiency evaluation}
To assess the efficiency of our method and verify its practical applicability, we count the number of model parameters and measure the average inference time required to process a single frame in the KITTI dataset. We conduct the same evaluation with other methods for comparison. Due to the out-of-memory issue encountered with AdaFit and GraphFit, we divided a single frame into 16 batches for all methods. As shown in Tab.~\ref{tab:table_efficiency}, our method demonstrates a significant advantage in terms of both efficiency and model size.

\subsubsection{Attention behavior visualisation}
A key aspect of our design involves establishing connections between semantically related regions, even if they are far away in space. This capability can be confirmed by visualising runtime attention maps generated by our trained model. Specifically, we create heatmaps utilising rows of matrix $\mat{A}$ in \eqref{eq:attn_norm} derived from the model’s inference process.

Fig.~\ref{fig:att_map} is produced using the synthetic dataset, showing that areas of interest are comprehensively considered for normal estimation at an individual position. Furthermore, this characteristic has been successfully generalised to the KITTI dataset without additional fine-tuning, as shown in Fig.~\ref{fig:att_map_kitti}. For instance, in Fig.~\ref{fig:att_map_kitti} (a), the distant points on the road aggregate information from numerous points across the entire road for geometric recovery, which is originally challenging due to the sparsity of the neighbouring LiDAR scans.

\section{Conclusion}
This paper presents a model that employs an attention mechanism for information fusion, with a specific application in estimating surface normals. We have also developed a comprehensive toolkit for data collection and a novel batching technique, both of which enable efficient and effective model training.

Multiple evaluations have been conducted, revealing that our model consistently delivers superior performance on both synthetic and real-world data. A limitation of \shortname\space is that the estimation is at where the LiDAR point is available. In future developments, it can be considered to predict surface normals and other geometric quantities at arbitrary positions.

\bibliographystyle{plain}
\bibliography{biblo.bib}

\begin{thebibliography}{10}

\bibitem{BansalRG16}
Aayush Bansal, Bryan~C. Russell, and Abhinav Gupta.
\newblock Marr revisited: 2d-3d alignment via surface normal prediction.
\newblock In {\em {CVPR}}, pages 5965--5974. {IEEE} Computer Society, 2016.

\bibitem{Ben-Shabat20-DeepFit}
Yizhak Ben{-}Shabat and Stephen Gould.
\newblock Deepfit: 3d surface fitting via neural network weighted least squares.
\newblock In Andrea Vedaldi, Horst Bischof, Thomas Brox, and Jan{-}Michael Frahm, editors, {\em {ECCV}}, volume 12346 of {\em Lecture Notes in Computer Science}, pages 20--34. Springer, 2020.

\bibitem{Ben-ShabatLF19}
Yizhak Ben{-}Shabat, Michael Lindenbaum, and Anath Fischer.
\newblock Nesti-net: Normal estimation for unstructured 3d point clouds using convolutional neural networks.
\newblock In {\em {CVPR}}, pages 10112--10120. Computer Vision Foundation / {IEEE}, 2019.

\bibitem{Vit}
Alexey Dosovitskiy, Lucas Beyer, Alexander Kolesnikov, Dirk Weissenborn, Xiaohua Zhai, Thomas Unterthiner, Mostafa Dehghani, Matthias Minderer, Georg Heigold, Sylvain Gelly, Jakob Uszkoreit, and Neil Houlsby.
\newblock An image is worth 16x16 words: Transformers for image recognition at scale.
\newblock In {\em 9th International Conference on Learning Representations, {ICLR} 2021, Virtual Event, Austria, May 3-7, 2021}. OpenReview.net, 2021.

\bibitem{Rethinking_normal_23}
Hang Du, Xuejun Yan, Jingjing Wang, Di~Xie, and Shiliang Pu.
\newblock Rethinking the approximation error in 3d surface fitting for point cloud normal estimation.
\newblock In {\em {IEEE/CVF} Conference on Computer Vision and Pattern Recognition, {CVPR} 2023, Vancouver, BC, Canada, June 17-24, 2023}, pages 9486--9495. {IEEE}, 2023.

\bibitem{EigenPF14}
David Eigen, Christian Puhrsch, and Rob Fergus.
\newblock Depth map prediction from a single image using a multi-scale deep network.
\newblock In {\em {NeurIPS}}, pages 2366--2374, 2014.

\bibitem{Rui2020}
Rui Fan, Hengli Wang, Bohuan Xue, Huaiyang Huang, Yuan Wang, Ming Liu, and Ioannis Pitas.
\newblock Three-filters-to-normal: An accurate and ultrafast surface normal estimator.
\newblock {\em {IEEE} Robotics Autom. Lett.}, 6(3):5405--5412, 2021.

\bibitem{FanelloVRKTDI17}
Sean~Ryan Fanello, Julien P.~C. Valentin, Christoph Rhemann, Adarsh Kowdle, Vladimir Tankovich, Philip~L. Davidson, and Shahram Izadi.
\newblock Ultrastereo: Efficient learning-based matching for active stereo systems.
\newblock In {\em {CVPR}}, pages 6535--6544. {IEEE} Computer Society, 2017.

\bibitem{GeigerLSU13}
Andreas Geiger, Philip Lenz, Christoph Stiller, and Raquel Urtasun.
\newblock Vision meets robotics: The {KITTI} dataset.
\newblock {\em Int. J. Robotics Res.}, 32(11):1231--1237, 2013.

\bibitem{GuennebaudG07}
Ga{\"{e}}l Guennebaud and Markus~H. Gross.
\newblock Algebraic point set surfaces.
\newblock {\em {ACM} Trans. Graph.}, 26(3):23, 2007.

\bibitem{GuerreroKOM18}
Paul Guerrero, Yanir Kleiman, Maks Ovsjanikov, and Niloy~J. Mitra.
\newblock Pcpnet learning local shape properties from raw point clouds.
\newblock {\em Comput. Graph. Forum}, 37(2):75--85, 2018.

\bibitem{tiv_GuoZC23}
Hongliang Guo, Jiankang Zhu, and Yunping Chen.
\newblock {E-LOAM:} lidar odometry and mapping with expanded local structural information.
\newblock {\em {IEEE} Trans. Intell. Veh.}, 8(2):1911--1921, 2023.

\bibitem{PCT}
Meng{-}Hao Guo, Junxiong Cai, Zheng{-}Ning Liu, Tai{-}Jiang Mu, Ralph~R. Martin, and Shi{-}Min Hu.
\newblock {PCT:} point cloud transformer.
\newblock {\em Comput. Vis. Media}, 7(2):187--199, 2021.

\bibitem{haas2014history}
John~K Haas.
\newblock A history of the unity game engine.
\newblock 2014.

\bibitem{HoppeDDMS92}
Hugues Hoppe, Tony DeRose, Tom Duchamp, John~Alan McDonald, and Werner Stuetzle.
\newblock Surface reconstruction from unorganized points.
\newblock In {\em {SIGGRAPH}}, pages 71--78. {ACM}, 1992.

\bibitem{Huang2023}
Jiahui Huang, Zan Gojcic, Matan Atzmon, Or~Litany, Sanja Fidler, and Francis Williams.
\newblock Neural kernel surface reconstruction.
\newblock In {\em {IEEE/CVF} Conference on Computer Vision and Pattern Recognition, {CVPR} 2023, Vancouver, BC, Canada, June 17-24, 2023}, pages 4369--4379. {IEEE}, 2023.

\bibitem{ml-agent}
Arthur Juliani, Vincent{-}Pierre Berges, Esh Vckay, Yuan Gao, Hunter Henry, Marwan Mattar, and Danny Lange.
\newblock Unity: {A} general platform for intelligent agents.
\newblock {\em CoRR}, abs/1809.02627, 2018.

\bibitem{KazhdanH13}
Michael~M. Kazhdan and Hugues Hoppe.
\newblock Screened poisson surface reconstruction.
\newblock {\em {ACM} Trans. Graph.}, 32(3):29:1--29:13, 2013.

\bibitem{KingmaB14}
Diederik~P. Kingma and Jimmy Ba.
\newblock Adam: {A} method for stochastic optimization.
\newblock In {\em {ICLR}}, 2015.

\bibitem{KrispelOWPB20}
Georg Krispel, Michael Opitz, Georg Waltner, Horst Possegger, and Horst Bischof.
\newblock Fuseseg: Lidar point cloud segmentation fusing multi-modal data.
\newblock In {\em {WACV}}, pages 1863--1872. {IEEE}, 2020.

\bibitem{LenssenOM20}
Jan~Eric Lenssen, Christian Osendorfer, and Jonathan Masci.
\newblock Deep iterative surface normal estimation.
\newblock In {\em {CVPR}}, pages 11244--11253. {IEEE}, 2020.

\bibitem{LiSDHH15}
Bo~Li, Chunhua Shen, Yuchao Dai, Anton van~den Hengel, and Mingyi He.
\newblock Depth and surface normal estimation from monocular images using regression on deep features and hierarchical crfs.
\newblock In {\em {CVPR}}, pages 1119--1127. {IEEE} Computer Society, 2015.

\bibitem{Li22-GraphFit}
Keqiang Li, Mingyang Zhao, Huaiyu Wu, Dong{-}Ming Yan, Zhen Shen, Fei{-}Yue Wang, and Gang Xiong.
\newblock Graphfit: Learning multi-scale graph-convolutional representation for point cloud normal estimation.
\newblock In Shai Avidan, Gabriel~J. Brostow, Moustapha Ciss{\'{e}}, Giovanni~Maria Farinella, and Tal Hassner, editors, {\em {ECCV}}, volume 13692, pages 651--667. Springer, 2022.

\bibitem{li2022hsurf}
Qing Li, Yu-Shen Liu, Jin-San Cheng, Cheng Wang, Yi~Fang, and Zhizhong Han.
\newblock {HSurf-Net}: Normal estimation for {3D} point clouds by learning hyper surfaces.
\newblock In {\em Advances in Neural Information Processing Systems (NeurIPS)}, volume~35, pages 4218--4230. Curran Associates, Inc., 2022.

\bibitem{RS-CNN}
Yongcheng Liu, Bin Fan, Shiming Xiang, and Chunhong Pan.
\newblock Relation-shape convolutional neural network for point cloud analysis.
\newblock In {\em {IEEE} Conference on Computer Vision and Pattern Recognition, {CVPR} 2019, Long Beach, CA, USA, June 16-20, 2019}, pages 8895--8904. Computer Vision Foundation / {IEEE}, 2019.

\bibitem{MerigotOG11}
Quentin M{\'{e}}rigot, Maks Ovsjanikov, and Leonidas~J. Guibas.
\newblock Voronoi-based curvature and feature estimation from point clouds.
\newblock {\em {IEEE} Trans. Vis. Comput. Graph.}, 17(6):743--756, 2011.

\bibitem{MitraNG04}
Niloy~J. Mitra, An~Thanh Nguyen, and Leonidas~J. Guibas.
\newblock Estimating surface normals in noisy point cloud data.
\newblock {\em Int. J. Comput. Geom. Appl.}, 14(4-5):261--276, 2004.

\bibitem{tiv_seg_peizhou}
Peizhou Ni, Xu~Li, Dong Kong, and Xiaoqing Yin.
\newblock Scene-adaptive 3d semantic segmentation based on multi-level boundary-semantic-enhancement for intelligent vehicles.
\newblock {\em IEEE Transactions on Intelligent Vehicles}, pages 1--11, 2023.

\bibitem{PyTorch}
Adam Paszke, Sam Gross, Francisco Massa, Adam Lerer, James Bradbury, Gregory Chanan, Trevor Killeen, Zeming Lin, Natalia Gimelshein, Luca Antiga, Alban Desmaison, Andreas K{\"{o}}pf, Edward~Z. Yang, Zachary DeVito, Martin Raison, Alykhan Tejani, Sasank Chilamkurthy, Benoit Steiner, Lu~Fang, Junjie Bai, and Soumith Chintala.
\newblock Pytorch: An imperative style, high-performance deep learning library.
\newblock In {\em NeurIPS}, pages 8024--8035. 2019.

\bibitem{PaulyKKG03}
Mark Pauly, Richard Keiser, Leif Kobbelt, and Markus~H. Gross.
\newblock Shape modeling with point-sampled geometry.
\newblock {\em {ACM} Trans. Graph.}, 22(3):641--650, 2003.

\bibitem{QiSMG17}
Charles~Ruizhongtai Qi, Hao Su, Kaichun Mo, and Leonidas~J. Guibas.
\newblock Pointnet: Deep learning on point sets for 3d classification and segmentation.
\newblock In {\em {CVPR}}, pages 77--85. {IEEE} Computer Society, 2017.

\bibitem{QiYSG17}
Charles~Ruizhongtai Qi, Li~Yi, Hao Su, and Leonidas~J. Guibas.
\newblock Pointnet++: Deep hierarchical feature learning on point sets in a metric space.
\newblock In {\em {NeurIPS}}, pages 5099--5108, 2017.

\bibitem{QiLLUJ22}
Xiaojuan Qi, Zhengzhe Liu, Renjie Liao, Philip H.~S. Torr, Raquel Urtasun, and Jiaya Jia.
\newblock Geonet++: Iterative geometric neural network with edge-aware refinement for joint depth and surface normal estimation.
\newblock {\em {IEEE} Trans. Pattern Anal. Mach. Intell.}, 44(2):969--984, 2022.

\bibitem{QiuCZZLZP19}
Jiaxiong Qiu, Zhaopeng Cui, Yinda Zhang, Xingdi Zhang, Shuaicheng Liu, Bing Zeng, and Marc Pollefeys.
\newblock Deeplidar: Deep surface normal guided depth prediction for outdoor scene from sparse lidar data and single color image.
\newblock In {\em {CVPR}}, pages 3313--3322. Computer Vision Foundation / {IEEE}, 2019.

\bibitem{RonnebergerFB15}
Olaf Ronneberger, Philipp Fischer, and Thomas Brox.
\newblock U-net: Convolutional networks for biomedical image segmentation.
\newblock In {\em {MICCAI}}, volume 9351 of {\em Lecture Notes in Computer Science}, pages 234--241. Springer, 2015.

\bibitem{tiv_SamalKSWM22}
Kruttidipta Samal, Hemant Kumawat, Priyabrata Saha, Marilyn Wolf, and Saibal Mukhopadhyay.
\newblock Task-driven rgb-lidar fusion for object tracking in resource-efficient autonomous system.
\newblock {\em {IEEE} Trans. Intell. Veh.}, 7(1):102--112, 2022.

\bibitem{how_bn}
Shibani Santurkar, Dimitris Tsipras, Andrew Ilyas, and Aleksander Madry.
\newblock How does batch normalization help optimization?
\newblock In {\em {NeurIPS}}, pages 2488--2498, 2018.

\bibitem{VizzoCCBS21}
Ignacio Vizzo, Xieyuanli Chen, Nived Chebrolu, Jens Behley, and Cyrill Stachniss.
\newblock Poisson surface reconstruction for lidar odometry and mapping.
\newblock In {\em {ICRA}}, pages 5624--5630. {IEEE}, 2021.

\bibitem{tiv_Li2023_multimodal_review}
Li~Wang, Xinyu Zhang, Ziying Song, Jiangfeng Bi, Guoxin Zhang, Haiyue Wei, Liyao Tang, Lei Yang, Jun Li, Caiyan Jia, and Lijun Zhao.
\newblock Multi-modal 3d object detection in autonomous driving: {A} survey and taxonomy.
\newblock {\em {IEEE} Trans. Intell. Veh.}, 8(7):3781--3798, 2023.

\bibitem{WuPCLZY21}
Zonghan Wu, Shirui Pan, Fengwen Chen, Guodong Long, Chengqi Zhang, and Philip~S. Yu.
\newblock A comprehensive survey on graph neural networks.
\newblock {\em {IEEE} Trans. Neural Networks Learn. Syst.}, 32(1):4--24, 2021.

\bibitem{CurveNet}
Tiange Xiang, Chaoyi Zhang, Yang Song, Jianhui Yu, and Weidong Cai.
\newblock Walk in the cloud: Learning curves for point clouds shape analysis.
\newblock In {\em 2021 {IEEE/CVF} International Conference on Computer Vision, {ICCV} 2021, Montreal, QC, Canada, October 10-17, 2021}, pages 895--904. {IEEE}, 2021.

\bibitem{spconv}
Yan Yan, Yuxing Mao, and Bo~Li.
\newblock {SECOND:} sparsely embedded convolutional detection.
\newblock {\em Sensors}, 18(10):3337, 2018.

\bibitem{ZengTHYSCW19}
Jin Zeng, Yanfeng Tong, Yunmu Huang, Qiong Yan, Wenxiu Sun, Jing Chen, and Yongtian Wang.
\newblock Deep surface normal estimation with hierarchical {RGB-D} fusion.
\newblock In {\em {CVPR}}, pages 6153--6162. Computer Vision Foundation / {IEEE}, 2019.

\bibitem{ZhangSYSLJF17}
Yinda Zhang, Shuran Song, Ersin Yumer, Manolis Savva, Joon{-}Young Lee, Hailin Jin, and Thomas~A. Funkhouser.
\newblock Physically-based rendering for indoor scene understanding using convolutional neural networks.
\newblock In {\em {CVPR}}, pages 5057--5065. {IEEE} Computer Society, 2017.

\bibitem{Refine-Net}
Haoran Zhou, Honghua Chen, Yingkui Zhang, Mingqiang Wei, Haoran Xie, Jun Wang, Tong Lu, Jing Qin, and Xiao{-}Ping Zhang.
\newblock Refine-net: Normal refinement neural network for noisy point clouds.
\newblock {\em {IEEE} Trans. Pattern Anal. Mach. Intell.}, 45(1):946--963, 2023.

\bibitem{Zhu21-AdaFit}
Runsong Zhu, Yuan Liu, Zhen Dong, Yuan Wang, Tengping Jiang, Wenping Wang, and Bisheng Yang.
\newblock Adafit: Rethinking learning-based normal estimation on point clouds.
\newblock In {\em {ICCV}}, pages 6098--6107. {IEEE}, 2021.

\end{thebibliography}

\begin{IEEEbiography}
[{\includegraphics[width=1in,height=1.25in,clip,keepaspectratio]{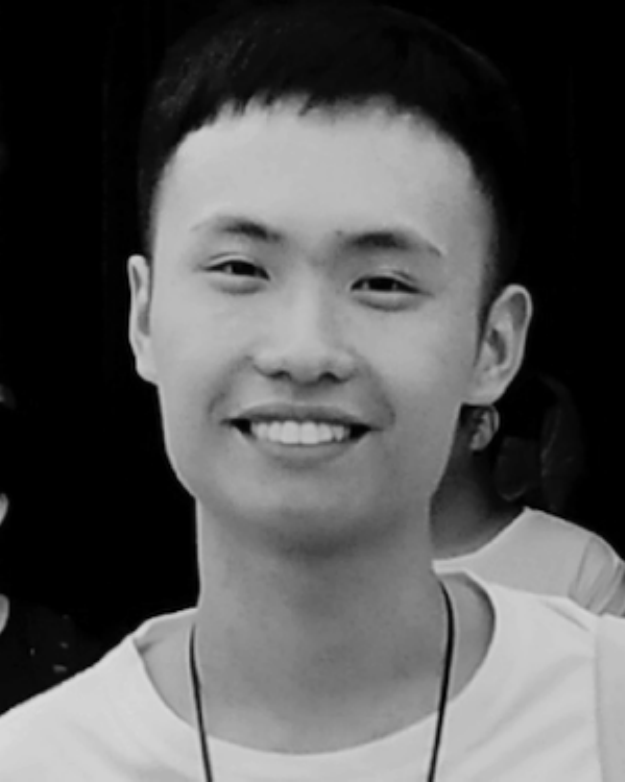}}]{Ancheng Lin}
received the B.Eng. degree in computer science from Guangdong Polytechnic Normal University, Guangdong, China, in 2019. He is currently pursuing the Ph.D. degree with the Faculty of Engineering and Information Technology, School of Computer Science, University of Technology Sydney, Australia. His research interests include 3D vision, autonomous driving, and deep learning.

\end{IEEEbiography}

\begin{IEEEbiography}
[{\includegraphics[width=1in,height=1.25in,clip,keepaspectratio]{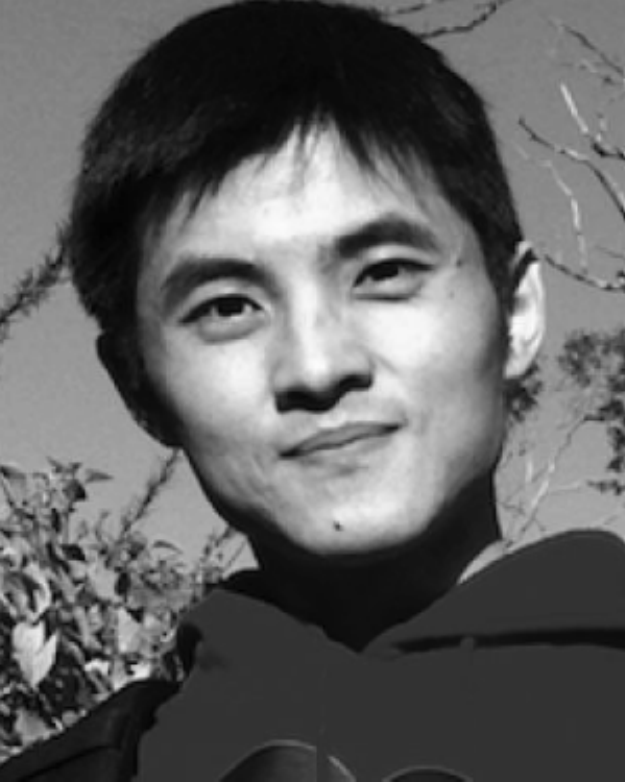}}]{Jun Li} received the B.S. degree in computer science and technology from Shandong University, China, in 2003, the M.Sc. degree in information and signal processing from Peking University, Beijing, China, in 2006, and the Ph.D. degree in computer science from Queen Mary University of London, U.K., in 2009. He is currently a Senior Lecturer with the Artificial Intelligence Institute, and the Faculty of Engineering and Information Technology, School of Computer Science, University of Technology Sydney, Australia. He also leads the AI Department, Elephant Tech LLC. His research interests include probabilistic data models and image and video analysis using neural networks.
\end{IEEEbiography}

\begin{IEEEbiography}
[{\includegraphics[width=1in,height=1.25in,clip,keepaspectratio]{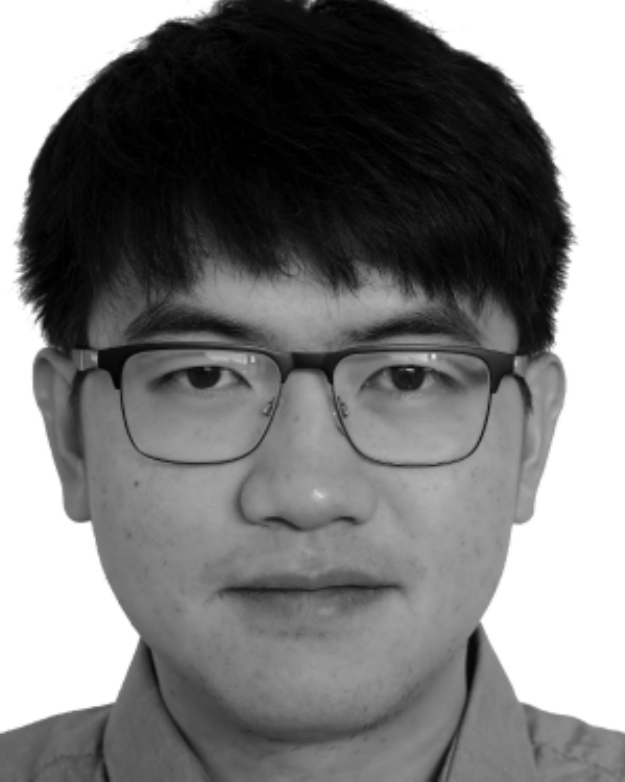}}]{Yusheng Xiang} (Member, IEEE) received the M.Sc. degree in vehicle engineering from the Karlsruhe Institute of Technology, Karlsruhe, Germany, in 2017, with a focus on mathematical model building and simulation, where he also got the Ph.D. degree from the Institute of Vehicle System Technology in 2021. He was a Research Scientist with Robert Bosch GmbH, Germany. From September 2020 to February 2021, he was a Visiting Scholar at the University of California at Berkeley, USA, supervised by Prof. Samuel S. Mao. He is also the CTO of Elephant Tech LLC, Shenzhen, China, a spinoff from Prof. Samuel S. Mao’s Laboratory. He has authored nine influential journals and international conference papers, and holds more than twenty patents. His group deals with improving mobile machines’ performance using artificial intelligence and the Internet of Things. 
\end{IEEEbiography}

\begin{IEEEbiography}
[{\includegraphics[width=1in,height=1.25in,clip,keepaspectratio]{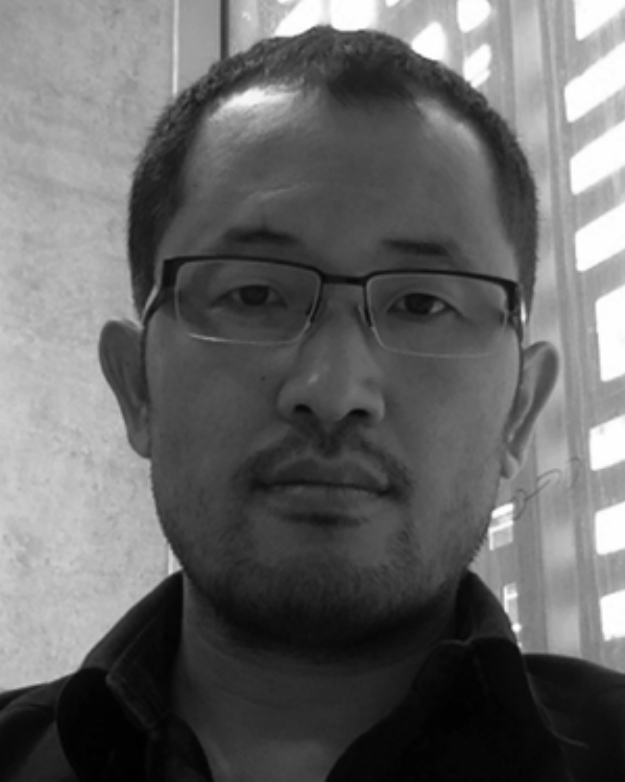}}]{Wei Bian}
received the B.Eng. degree in electronic engineering, the B.Sc. degree in applied mathematics, and the M.Eng. degree in electronic engineering from the Harbin Institute of Technology, Harbin, China, in 2005 and 2007, respectively, and the Ph.D. degree in computer science from the University of Technology Sydney, Australia, in 2012. He is a currently a Lecturer with the University of Technology Sydney. His research interests include machine learning and computer vision.
\end{IEEEbiography}

\begin{IEEEbiography}
[{\includegraphics[width=1in,height=1.25in,clip,keepaspectratio]{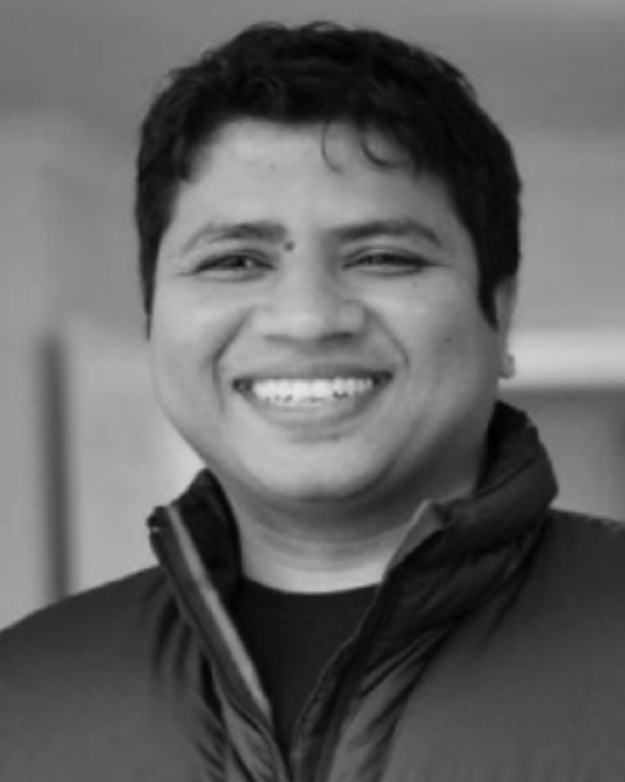}}]{Mukesh Prasad}
(Senior Member, IEEE) is a Senior Lecturer at the School of Computer Science in the Faculty of Engineering and IT at UTS who has made substantial contributions to the fields of machine learning, artificial intelligence and Natural Language Processing.  He is working also in the evolving and increasingly important field of image processing, data analytics and edge computing, which promise to pave the way for the evolution of new applications and services in the areas of healthcare, biomedical, agriculture, smart cities, education, marketing and finance. His research has appeared in numerous prestigious journals, including IEEE/ACM Transactions, and at conferences, and he has written more than 150 research papers. He started his academic career as a lecturer with UTS in 2017 and became a core member of the University’s world-leading Australian Artificial Intelligence Institute (AAII), which has a vision to develop theoretical foundations and advanced technologies for AI and to drive progress in related areas. His research is backed by industry experience, specifically in Taiwan, where he was the principal engineer (2016-17) at the Taiwan Semiconductor Manufacturing Company (TSMC). There, he developed new algorithms for image processing and pattern recognition using machine learning techniques. He received an M.S. degree from the School of Computer and Systems Sciences at the Jawaharlal Nehru University in New Delhi, India (2009), and a PhD from the Department of Computer Science at the National Chiao Tung University in Taiwan (2015).
\end{IEEEbiography}

\end{document}